\documentclass[10pt]{article}

\usepackage[letterpaper, margin=1in]{geometry}
\usepackage{iftex, amsmath}
\ifPDFTeX                                     
  \usepackage[T1]{fontenc}
  \usepackage{libertinus}                     
  \usepackage{libertinust1math}               
  \usepackage[scaled=0.9]{zi4}                
\else                                         
  \usepackage{libertinus}                     
\fi
\usepackage[final]{microtype}
\usepackage{xcolor}
\definecolor{encoder}{HTML}{2A78D6}           
\definecolor{decoder}{HTML}{EB6834}           
\definecolor{ink2}{HTML}{52514E}
\usepackage[colorlinks, linkcolor=encoder, citecolor=encoder, urlcolor=encoder]{hyperref}
\usepackage[capitalise, noabbrev]{cleveref}
\usepackage[numbers, sort&compress]{natbib}

\usepackage[sf, bf, small, compact]{titlesec}
\titlespacing*{\section}{0pt}{1.4em}{0.5em}
\titlespacing*{\subsection}{0pt}{1em}{0.3em}
\usepackage[font=small, labelfont={sf, bf}, labelsep=period, skip=5pt]{caption}

\usepackage{booktabs, graphicx}
\usepackage[section]{placeins}   
\usepackage{tikz}
\usetikzlibrary{positioning, arrows.meta, fit, calc}
\graphicspath{{generated/}}
\newcommand{\TestUbYesNo}{71.3}

\newcommand{\ExtUbYesNo}{59.6}
\newcommand{\TestUbSqrt}{71.2}

\newcommand{\TestUbSig}{72.1}

\newcommand{\TestUbList}{69.7}

\newcommand{\ExtUbList}{57.8}
\newcommand{\TestQwFull}{67.6}

\newcommand{\ExtQwFull}{56.1}
\newcommand{\TestQwLora}{63.0}

\newcommand{\ExtQwLora}{56.9}
\newcommand{\TestQwFourLora}{77.3}

\newcommand{\DiffUbYesNoUbListAbs}{1.6}

\newcommand{\DiffUbSigUbYesNoCI}{[+0.2, +1.3]}
\newcommand{\DiffUbSigUbYesNoAbs}{0.8}

\newcommand{\DiffQwFullQwLoraAbs}{4.6}

\newcommand{\EvSentAPList}{0.55}
\newcommand{\EvTokAPList}{0.39}

\newcommand{\EvMsList}{68}
\newcommand{\EvThrList}{0.74}

\newcommand{\EvSentAPGold}{0.53}
\newcommand{\EvTokAPGold}{0.38}

\newcommand{\EvSentAPSig}{0.68}
\newcommand{\EvTokAPSig}{0.46}

\newcommand{\EvPTenSig}{0.63}
\newcommand{\EvChanceSig}{0.17}

\newcommand{\EvThrSig}{0.79}
\newcommand{\EvWordPSig}{0.33}
\newcommand{\EvWordRSig}{0.35}
\newcommand{\EvRows}{889}
\newcommand{\EvTuneRows}{267}
\newcommand{\AtTwoSentAP}{0.33}

\newcommand{\ContextCiteSentAP}{0.41}

\newcommand{\ContextCiteLds}{0.64}
\newcommand{\ContextCiteMs}{1{,}860}
\newcommand{\RpsUb}{75.5}
\newcommand{\RpsQwFull}{16.4}
\newcommand{\RpsQwFourLora}{6.5}

\newcommand{\MemPeakUb}{2.8}

\newcommand{\MsUb}{29}

\newcommand{\MemPeakUbBf}{1.9}

\newcommand{\MemPeakQwFull}{2.2}

\newcommand{\MsQwFull}{68}
\newcommand{\MsCopyQwFull}{92}
\newcommand{\MemPeakCopyQwFull}{8.6}
\newcommand{\MsPreQwFull}{36}

\newcommand{\MsQwFourLora}{124}

\newcommand{\MsPreQwFourLora}{85}
\newcommand{\NTrain}{13{,}000}
\newcommand{\NHeldOut}{6{,}315}
\newcommand{\NDev}{1{,}300}
\newcommand{\MaxTokens}{10{,}240}
\newcommand{\NPerDataset}{1{,}000}
\newcommand{\NTest}{5{,}015}

\newcommand{\CpExtYes}{57.8}
\newcommand{\CpShufYes}{15.6}
\newcommand{\CpDropYes}{5.9}
\newcommand{\CpGradYes}{842}
\newcommand{\CpLabelYes}{0.167}

\newcommand{\CpAccYesNo}{71.5}
\newcommand{\CpExtYesNo}{59.6}
\newcommand{\CpShufYesNo}{14.6}
\newcommand{\CpDropYesNo}{4.8}
\newcommand{\CpGradYesNo}{139}
\newcommand{\CpLabelYesNo}{0.213}

\newcommand{\CpEceYesNo}{0.027}

\newcommand{\CpShufSqrt}{16.6}

\newcommand{\CpGradSqrt}{104}

\newcommand{\CpAccSig}{72.9}

\newcommand{\CpShufSig}{13.7}

\newcommand{\CpLabelSig}{0.139}

\newcommand{\CpEceSig}{0.023}

\newcommand{\CpSigGoldSig}{0.049}

\title{\sffamily\bfseries\Large
  UltraBERT-JEV: Encoder models for System One Applications}
\author{\sffamily
  Astarag Mohapatra\\[0.2em]
  {\small\color{ink2} Independent Researcher}\\[0.2em]
  {\small\color{ink2} \href{mailto:astarag2843@gmail.com}{astarag2843@gmail.com}}}
\date{\small\sffamily\color{ink2}\today}

\begin{document}
\maketitle

\begin{abstract}
We study encoder models that pick one of several options in a single forward pass, without generating text.
This technical report compares recipes for fine-tuning UltraBERT-380M \citep{ultrabert} with one mask per option.
The four recipes differ in how an option is scored and how the scores become probabilities; one follows the expert gate of DeepSeek-V3 \citep{deepseekv3}.
Trained on the same data as Qwen3-0.6B with the LLM-as-Jev recipe \citep{li2026llmasjev} (\textbf{\TestQwFull{}\%} test accuracy), the encoder reaches \textbf{\TestUbList{}\%} to \textbf{\TestUbSig{}\%}.
Scoring an option by ``yes'' minus ``no'' instead of ``yes'' alone gives the larger gain, from \TestUbList{}\% to \TestUbYesNo{}\% on the test set and from \ExtUbList{}\% to \textbf{\ExtUbYesNo{}\%} on external benchmarks (\textbf{\ExtQwFull{}\%} for the decoder).
Sigmoid normalisation adds \DiffUbSigUbYesNoAbs{} test points; a square-root softplus adds none.
One request at a time on an NVIDIA A100, the encoder takes \textbf{\MsUb{}\,ms} per decision against \textbf{\MsQwFull{}\,ms}, but uses more peak memory (\MemPeakUb{} against \MemPeakQwFull{}\,GiB).
An auxiliary head, trained on human-marked evidence and on labels selected with a reward adapted from SelfCite \citep{selfcite}, predicts in the same pass the sentences annotated as support for the answer.
\end{abstract}

\section{Introduction}
\label{sec:intro}

Software often has to choose among fixed options: route a support ticket, classify an intent, or check a claim against evidence.
A chat model can do this, but its answer must be generated and parsed.
Jev-style decision models \citep{jev} instead return a probability distribution over the options, so a new task needs only a prompt and a list of options.
LLM-as-Jev \citep{li2026llmasjev} shows that an unmodified decoder language model works as such a model: it reads numbered options and answers with a number.
Fine-tuning it with KL anchors helps most for small models and for tasks with many options.

A decision needs almost no generation, while decoders are built for it.
Multi-query and grouped-query attention \citep{mqa, gqa} shrink the KV cache for decoding, and the parameters they save can go into larger MLP layers.
\citet{prefillonly} show that these layers cause memory spikes in workloads that only read their input.
An encoder keeps no KV cache, which is one reason to try it; at our scale, \cref{sec:cost} finds no peak-memory advantage.

An encoder also reads the input in both directions.
A classifier head would fix the label set, but a masked language model can score any option by whether it fills a mask next to the option with ``yes'' or ``no''.
Recent encoders such as ModernBERT \citep{modernbert} and UltraBERT \citep{ultrabert} read the long inputs that decisions over documents need.

Reading in both directions also helps with evidence.
A decoder serving many questions about one document usually reads the document first, so that its KV cache can be reused, as with prefix caching in vLLM \citep{vllm} and RadixAttention in SGLang \citep{sglang}.
The cached document states then do not depend on the question.
Our encoder reads the question and the document together.

Our contributions are:
\begin{itemize}\setlength{\itemsep}{2pt}\setlength{\topsep}{2pt}
  \item \textbf{An encoder decision model.}
    UltraBERT-380M with \emph{one mask per option}, scoring \emph{every option in one forward pass} with \emph{no new head}, for any number of options.
    It is fine-tuned on the LLM-as-Jev training mixture with Nimble added (\cref{sec:data}).
  \item \textbf{Training recipes.}
    \emph{Four ways to score and normalise} the mask predictions, one following the DeepSeek-V3 expert gate \citep{deepseekv3}.
    \emph{Scoring by ``yes'' minus ``no''} gives the larger gain; \emph{sigmoid normalisation} adds \DiffUbSigUbYesNoAbs{} test points, mostly on MuSiQue.
  \item \textbf{A comparison with decoders under matched data.}
    Against Qwen3-0.6B \citep{qwen3} fully fine-tuned on the same examples, \emph{every encoder recipe is more accurate} (\TestUbList{}\% to \TestUbSig{}\% against \TestQwFull{}\% on a held-out test split) and \emph{faster} (\MsUb{} against \MsQwFull{}\,ms per decision), but \emph{not lighter} in peak memory.
    Two reproduced LoRA models \citep{lora} serve as references.
  \item \textbf{Predicting supporting evidence.}
    An \emph{auxiliary evidence head} in the same forward pass, trained on human-marked labels and labels selected with a reward adapted from SelfCite \citep{selfcite}.
    On the softmax encoder it \emph{ranks annotated sentences higher than ContextCite and AT2} \citep{contextcite, at2}.
\end{itemize}

\section{Method}
\label{sec:method}

\subsection{Input format}
\label{sec:format}

\begin{figure}[t]
  \centering
  \begin{tikzpicture}[x=1in, y=1in,
      font=\sffamily\scriptsize,
      prompt/.style={draw=ink2, line width=0.4pt, rounded corners=2pt, fill=black!3, inner sep=4pt,
                     align=left, font=\ttfamily\scriptsize},
      box/.style={draw, rounded corners=2pt, inner sep=4pt, align=center, text width=2.85in},
      arr/.style={-{Stealth[length=4pt]}, line width=0.6pt, color=ink2},
      head/.style={font=\sffamily\bfseries\small, anchor=west}]
    \node[head, text=decoder] at (0, 0) {LLM-as-Jev: decoder};
    \node[prompt, text width=2.85in, anchor=north west] (dp) at (0, -0.15) {%
      State: What size does the screen need to be cut?\\
      Which solution better achieves the goal?\\
      Options:\\
      \relax[1] Cut to the size of the jar opening.\\
      \relax[2] Long enough to cover the whole jar.\\
      Select one option. Answer only with its number.\\
      \textbf{Best answer:} \relax[\,\textcolor{decoder}{\textbf{1\,]}}};
    \node[box, draw=decoder, below=0.22in of dp] (dt) {Loss on the \textbf{digits of the option number}\\
      (a trie), plus KL anchors to the base model};
    \node[box, draw=decoder, below=0.2in of dt] (dd) {Decision: the most probable option number};
    \draw[arr] (dp) -- (dt);  \draw[arr] (dt) -- (dd);
    \node[head, text=encoder] at (3.45, 0) {UltraBERT: encoder (this work)};
    \node[prompt, text width=2.2in, anchor=north west] (ep) at (3.45, -0.15) {%
      State: What size does the screen need to be cut?\\
      Which solution better achieves the goal?\\
      Mark the correct option with yes\\and every other option with no.\\
      Options:\\
      \relax[1] \colorbox{encoder!18}{[M]} Cut to the size of the jar opening.\\
      \relax[2] \colorbox{encoder!18}{[M]} Long enough to cover the whole jar.};
    \node[align=left, text=encoder] (s1) at ($(ep.south east)+(0.42in, 0.31in)$) {$s_1$};
    \node[align=left, text=encoder] (s2) at ($(ep.south east)+(0.42in, 0.14in)$) {$s_2$};
    \draw[arr, color=encoder] ($(ep.south east)+(0, 0.31in)$) -- (s1);
    \draw[arr, color=encoder] ($(ep.south east)+(0, 0.14in)$) -- (s2);
    \node[box, draw=encoder, text width=1.28in, anchor=north west] (sm) at ($(ep.south west)+(0,-0.22in)$) {%
      \textbf{Plain softmax}\\[1pt]
      $s_k=\text{logit}(\textit{yes})$\\[2pt]
      $p_k=\dfrac{e^{s_k}}{\sum_j e^{s_j}}$\\[3pt]
      loss $=-\log p_{\text{gold}}$};
    \node[box, draw=encoder, fill=encoder!8, text width=1.28in, anchor=north west] (sg) at ($(sm.north east)+(0.14in,0)$) {%
      \textbf{Sigmoid-norm}\\[1pt]
      $s_k=\text{logit}(\textit{yes})-\text{logit}(\textit{no})$\\[2pt]
      $p_k=\dfrac{\sigma(s_k)}{\sum_j\sigma(s_j)}$\\[3pt]
      loss $=-\log p_{\text{gold}}$};
    \node[box, draw=encoder, text width=2.7in, anchor=north west] (ed) at ($(sm.south west)+(0,-0.2in)$) {Decision: $\arg\max_k s_k$\\one forward pass, no generation};
    \draw[arr] ($(sm.north)+(0,0.22in)$) -- (sm.north);
    \draw[arr] ($(sg.north)+(0,0.22in)$) -- (sg.north);
    \draw[arr] (sm.south) -- (sm.south |- ed.north);
    \draw[arr] (sg.south) -- (sg.south |- ed.north);
  \end{tikzpicture}
  \caption{One row, two ways. The decoder reads the numbered options and writes the number of its answer; the
  loss is taken over the tokens of that number. The encoder gets a mask in front of each option and scores each
  option directly. Two of the four recipes are shown: the plain softmax over the options, which scores an
  option by the ``yes'' logit alone, and the sigmoid-normalised loss, which scores it by ``yes'' minus ``no''.
  Each option's score goes through a sigmoid, and the sum in the denominator keeps the options competing. The other two normalise ``yes'' minus ``no'' with a softmax and with a square-root softplus.}
  \label{fig:intro}
\end{figure}
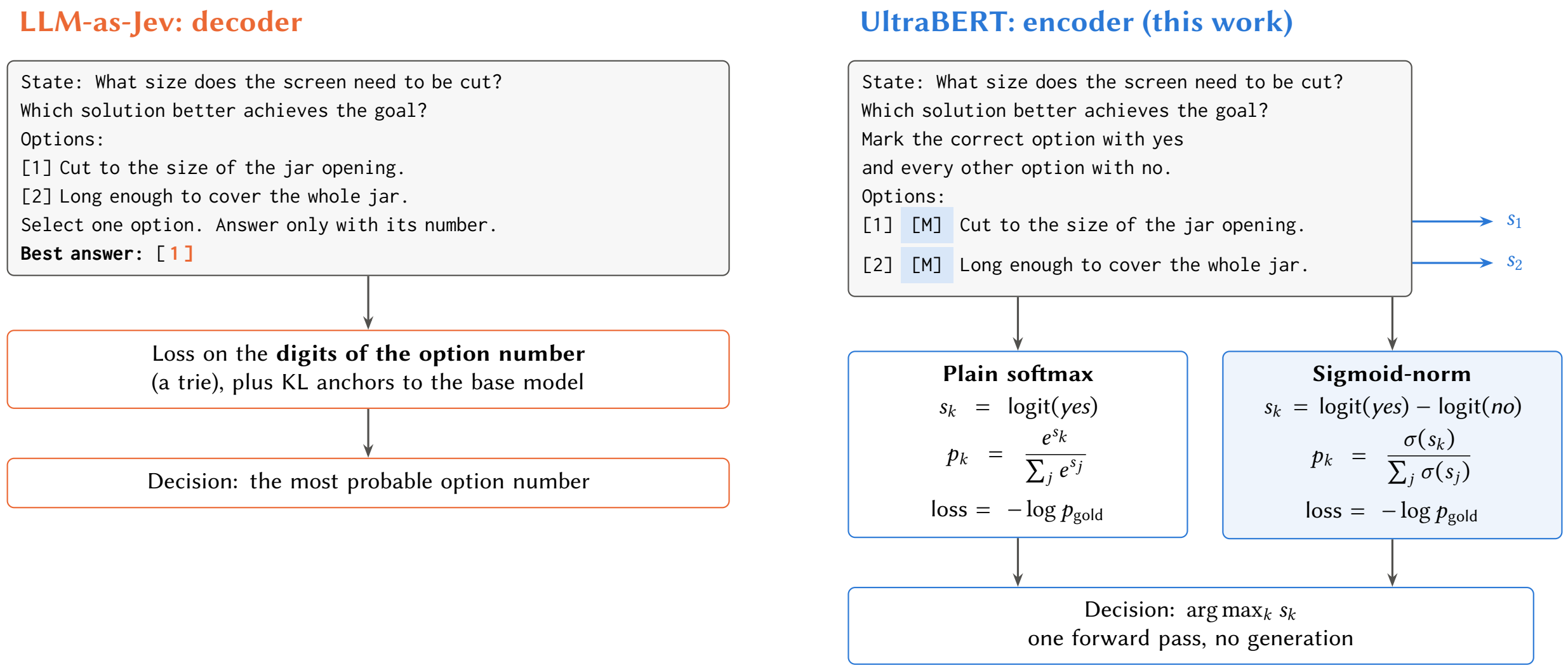

\Cref{fig:intro} shows one row in both formats.
The options are shuffled each time a row is used, so the position of the correct option carries no information.

\paragraph{Decoder.}
The prompt follows LLM-as-Jev: the state, the question, the options numbered $[1]$ to $[K]$, and an instruction to answer only with the number.
The reply is filled in up to ``Best answer: ['', with thinking turned off, so each option $k$ is the token sequence ``$k$]'' and the options form a trie.
The decoder chooses the option whose whole sequence has the highest log-probability.

\paragraph{Encoder.}
The state and question are the same, the instruction asks to mark the correct option with ``yes'' and the others with ``no'', and each option is listed as ``$[k]$ \texttt{[M]} description'', with a mask token in front.
There is no chat template.
One forward pass gives the logits at every mask, and option $k$ is scored from the logits of ``yes'' and ``no'', each with a leading space, at its mask.

\subsection{Loss functions}
\label{sec:loss}

\paragraph{Decoder: trie loss with KL anchors.}
The tree loss is the negative log-probability of the gold option's path through the trie, each step normalised over the tokens legal there,
\begin{equation}
  \mathcal{L}_{\text{tree}} = -\sum_{t}\log\frac{\exp z_t(g_t)}{\sum_{v\in\mathcal{V}_t}\exp z_t(v)},
\end{equation}
where $z_t$ are the logits at step $t$, $g_t$ the gold token and $\mathcal{V}_t$ the legal tokens.
Three anchor terms keep the model close to its starting version $p_0$ \citep{li2026llmasjev}: $\mathcal{L}_{\text{mass}}$ keeps the total probability on the legal tokens, $\mathcal{L}_{\text{out}}$ keeps how the remaining probability is spread over the rest of the vocabulary, and $\mathcal{L}_{\text{pos}}$ keeps the next-token distribution at the start of the reply and after the closing bracket.
They are KL divergences to $p_0$, computed over its 64 most likely tokens and one bucket for the rest.
The loss is $\mathcal{L}=\mathcal{L}_{\text{tree}}+\lambda\,(\mathcal{L}_{\text{mass}}+\mathcal{L}_{\text{out}}+\mathcal{L}_{\text{pos}})$ with $\lambda=1$.

\paragraph{Encoder recipes.}
Let $\ell_k(v)$ be the logit of token $v$ at mask $k$, and let $g$ be the gold option.
A score $s_k$ says how strongly the model accepts option $k$; the decision is the option with the highest score.
Training raises the gold option's score relative to the other options, but no mask gets its own yes-or-no target.
Thus, a chosen option can still prefer ``no'': for example, scores of $-3$ and $-8$ select the first option.
We train the following four recipes:
\begin{itemize}
  \item \textbf{Yes-logit softmax.}
    The score is the ``yes'' logit alone, and the options compete through one softmax:
    \begin{equation}
      s_k=\ell_k(\textit{yes}),\qquad
      p_k=\frac{e^{s_k}}{\sum_{j=1}^{K}e^{s_j}},\qquad
      \mathcal{L}_{\text{softmax}}=-\log p_g.
    \end{equation}

  \item \textbf{Yes-minus-no softmax.}
    The score measures the preference for ``yes'' over ``no'', followed by the same softmax:
    \begin{equation}
      s_k=\ell_k(\textit{yes})-\ell_k(\textit{no}),\qquad
      p_k=\frac{e^{s_k}}{\sum_{j=1}^{K}e^{s_j}},\qquad
      \mathcal{L}_{\text{softmax}}=-\log p_g.
    \end{equation}

  \item \textbf{Yes-minus-no sigmoid normalisation.}
    Each score first goes through a sigmoid, and the resulting values are normalised by their sum:
    \begin{equation}
      s_k=\ell_k(\textit{yes})-\ell_k(\textit{no}),\qquad
      p_k=\frac{\sigma(s_k)}{\sum_{j=1}^{K}\sigma(s_j)},\qquad
      \mathcal{L}_{\text{sig-norm}}=-\log p_g.
    \end{equation}
    This is the form of the DeepSeek-V3 expert gate \citep{deepseekv3}, with options in place of experts, normalised over all $K$ options and with no selection or balancing bias.
    Each $\sigma(s_k)$ lies between 0 and 1, but only the ratios enter $p_k$, so the options still compete; the decision is still the highest score.
    The gradient with respect to $s_k$ is
    \begin{equation}
      \left(1-\sigma(s_k)\right)
      \left(p_k-\mathbf{1}[k=g]\right).
    \end{equation}
    When an option's sigmoid approaches 1, the direct gradient of the loss with respect to its score becomes small.
    We exclude the sigmoid on the ``yes'' logit alone: vocabulary logits are around 200 (\cref{sec:precision}), so every sigmoid would saturate near 1 and the gradient would be negligible.

  \item \textbf{Yes-minus-no square-root softplus.}
    As an unbounded alternative to the sigmoid, we use
    $h(s)=\sqrt{\log(1+e^s)}$:
    \begin{equation}
      s_k=\ell_k(\textit{yes})-\ell_k(\textit{no}),\qquad
      p_k=\frac{h(s_k)}{\sum_{j=1}^{K}h(s_j)},\qquad
      \mathcal{L}_{\text{sqrt-sp}}=-\log p_g.
    \end{equation}
    For a clearly rejected option ($s\ll0$), $h(s)\approx e^{s/2}$, so the options behave as in a softmax at twice the temperature.
    For a clearly accepted option ($s\gg0$), $h(s)\approx\sqrt{s}$ keeps growing, whereas the sigmoid stops at 1.
    The direct gradient with respect to its score therefore shrinks slowly, where the sigmoid's nearly vanishes.
\end{itemize}

\section{Experimental setup}
\label{sec:setup}

\subsection{Training data}
\label{sec:data}

We train on the Reason+Long mixture of LLM-as-Jev \citep{li2026llmasjev}, with Nimble added: 13 datasets covering intent routing (CLINC150 \citep{clinc150}, MASSIVE \citep{massive}, Bitext \citep{bitext}), short reasoning (ReClor \citep{reclor}, LogiQA~2.0 \citep{logiqa2}, CosmosQA \citep{cosmosqa}, PIQA \citep{piqa}, ANLI \citep{anli}, ProofWriter \citep{proofwriter}), long input (ContractNLI \citep{contractnli}, MuSiQue \citep{musique}, QuALITY \citep{quality}) and Nimble \citep{nimble}.
Each row is a state (the passages), a question and a list of options.
For example, a MuSiQue row gives about twenty paragraphs and a multi-hop question, and asks whether the paragraphs contain enough information to answer it (``Yes'' or ``No'').

We keep \NPerDataset{} training rows per dataset, \NTrain{} in total, after these filters:
\begin{enumerate}\setlength{\itemsep}{0pt}\setlength{\topsep}{2pt}
  \item \textbf{Duplicates.} Rows with the same input are collapsed, and removed if their answers differ.
  \item \textbf{Leakage.} A training row is removed if its input or source document appears in the validation split.
  \item \textbf{Benchmark overlap.} A training row is removed if it shares a 13-word n-gram, or a sentence of six or more words, with an evaluation benchmark.
  \item \textbf{Length.} Prompts over \MaxTokens{} tokens (Qwen3.5-4B tokenizer) are dropped, never truncated.
\end{enumerate}

The held-out rows come from the validation split, with the same duplicate and length filters, up to 500 per dataset and in whole source documents (\NHeldOut{} rows).
A random 100 rows per dataset form the dev set (\NDev{} rows), used to choose learning rates and the evidence threshold; the other \NTest{} rows form the test set, excluded from model selection.
The split is by row, so a dev and a test row can share a source document, such as two statements about one contract.

\subsection{Models}
\label{sec:models}

UltraBERT-380M \citep{ultrabert} is fully fine-tuned with each of the four recipes of \cref{sec:loss}.
Qwen3-0.6B \citep{qwen3} is fully fine-tuned with the decoder loss, with a frozen copy as $p_0$; it is the matched comparison.
As references, we also report two LoRA models \citep{lora}, Qwen3-0.6B and Qwen3-4B, reproduced from LLM-as-Jev on the same training set and loss.

\subsection{Implementation details}
\label{sec:Implementation details}

Every model is trained for one epoch with a batch of 32 rows.
We sweep the learning rate on dev, one run per value: $\{0.8, 1.5, 3, 5\}\times10^{-5}$ for the encoder with the softmax over ``yes'', $\{0.8, 1.5, 2, 3\}\times10^{-5}$ with the sigmoid-normalised loss, and $\{0.5, 1, 2\}\times10^{-5}$ for the decoder.
Both encoder sweeps choose $1.5\times10^{-5}$, which we reuse for the other two recipes; for Qwen the two best rates differ by 0.1 points, and we use $1\times10^{-5}$.
At the chosen rate we train each model several times, changing only the row order and option shuffling, and report the average.

The metric is accuracy, and ALL is its mean over the 13 datasets.
On the test set we average each row's correctness over a model's runs before computing ALL.
For a difference between two models we give a paired 95\% bootstrap interval: 2{,}000 resamples of the test rows within each dataset, with both models scored on the same rows.
It covers test-row sampling only, not retraining, and treats rows from one document as independent, so it can be too narrow.

The encoder is scored with fp32 weights under bf16 autocast, since bf16 weights change its long-document results (\cref{sec:precision}); the Qwen checkpoints use bf16 weights.

\subsection{External benchmarks and cost}
\label{sec:external}

To see how training changes behaviour outside the mixture, we score six benchmarks as decision rows, never used for any choice: BoolQ \citep{boolq}, MMLU \citep{mmlu}, MMLU-Pro \citep{mmlupro}, ARC-Challenge \citep{arc}, Winogrande \citep{winogrande} and SciQ \citep{sciq}.
We draw fixed random samples of 500, 1200, 800, 600, 500 and 400 rows, the sizes of LLM-as-Jev, whose samples are not released.
External is the mean accuracy over the six.

Time per decision is the mean latency of one request over the dev rows, scored one at a time on one A100, as when a pipeline sends each item as it arrives.
It covers the forward pass and option scoring but not tokenisation, after 20 warm-up rows, with the GPU synchronised around every call.
The dev rows average 1{,}100 to 1{,}200 tokens.
The encoder is timed on the softmax checkpoint and the decoders on the reproduced LoRA checkpoints with merged weights; the other checkpoints of each model share their architecture.
\Cref{app:impl} gives batched throughput.

\section{Results}
\label{sec:results}

\subsection{Encoder against decoder}

\Cref{tab:headline} gives the main comparison.
Every encoder recipe is above fully fine-tuned Qwen3-0.6B, both on the test set (\TestUbList{}\% to \TestUbSig{}\% against \TestQwFull{}\%) and on the external benchmarks (\ExtUbList{}\% to \ExtUbYesNo{}\% against \ExtQwFull{}\%), with fewer parameters (380M against 0.6B).
These results concern these two models under our setup; the reproduced Qwen3-4B LoRA model, at \TestQwFourLora{}\%, remains more accurate.

\begin{table}[t]
  \centering\small
  \caption{Matched comparison: average accuracy (\%) over training runs. Test ALL is on the test set;
  Dev ALL and External are on the dev set and the six external benchmarks. Learning rates were chosen on
  dev. Bold: best below 1B parameters.}
  \label{tab:headline}
\begin{tabular}{@{}llccc@{}}
\toprule
Model & Params & Dev ALL & Test ALL & External \\
\midrule
UltraBERT-380M, softmax, yes & 380M & 71.2 & 69.7 & 57.8 \\
UltraBERT-380M, softmax, yes $-$ no & 380M & 71.7 & 71.3 & \textbf{59.6} \\
UltraBERT-380M, sqrt-softplus, yes $-$ no & 380M & 71.6 & 71.2 & 59.3 \\
UltraBERT-380M, sigmoid-norm, yes $-$ no & 380M & \textbf{72.7} & \textbf{72.1} & 59.5 \\
Qwen3-0.6B, full FT & 0.6B & 69.0 & 67.6 & 56.1 \\
Qwen3-0.6B, LoRA (reproduced) & 0.6B & 64.5 & 63.0 & 56.9 \\
\midrule
Qwen3-4B, LoRA (reproduced) & 4B & 79.3 & 77.3 & 73.5 \\
\bottomrule
\end{tabular}

\end{table}

\subsection{Full fine-tuning against LoRA}

The reproduced Qwen3-0.6B LoRA model reaches \TestQwLora{}\% on the test set, \DiffQwFullQwLoraAbs{} points below full fine-tuning, so full fine-tuning is the stronger baseline.
Its external accuracy (\ExtQwLora{}\%) is not below that of the full fine-tune (\ExtQwFull{}\%).

\subsection{Recipes}

Changing the score from ``yes'' to ``yes'' minus ``no'', both with the softmax, adds \DiffUbYesNoUbListAbs{} test points and raises external accuracy from \ExtUbList{}\% to \ExtUbYesNo{}\%.
Sigmoid normalisation adds a further \DiffUbSigUbYesNoAbs{} test points (95\% interval \DiffUbSigUbYesNoCI{}, \cref{sec:Implementation details}) and nothing on the external benchmarks.

This gain is concentrated in MuSiQue (69.8\% against 63.0\%), which contributes about 0.5 of the \DiffUbSigUbYesNoAbs{} points.
On the other twelve datasets it averages 0.3 points and goes both ways, from $+3.0$ on Nimble to $-2.4$ on PIQA (\cref{tab:perdataset}).

The square-root softplus, which has no upper bound, does not reproduce the gain (\TestUbSqrt{}\% against \TestUbYesNo{}\% for the softmax).
The comparison does not show why: the two functions also differ in their gradients, and for rejected options the square-root softplus acts as a softmax at twice the temperature.

\subsection{Sensitivity to option order and removal}
\label{sec:coupling}

On the dev rows we score each row in four random option orders and once with one wrong option removed, and count how often the answer changes (\cref{tab:coupling}).
With the raw scores fixed, no normalisation could change the winner, so a change means the encoder recomputed its scores after reading the new list: the test measures the trained model.
The two softmax recipes isolate the score; the softmax and the sigmoid over ``yes'' minus ``no'' isolate the normalisation used in training.

The score does most of the work.
``Yes'' minus ``no'' raises external accuracy from \CpExtYes{}\% to \CpExtYesNo{}\%, lowers the answer changes on removal from \CpDropYes{}\% to \CpDropYesNo{}\%, and lowers the largest gradient norm from \CpGradYes{} to \CpGradYesNo{}.
Under reordering, the score and the sigmoid each remove about one point of answer changes (\CpShufYes{}\%, \CpShufYesNo{}\%, \CpShufSig{}\%).

On top of the score, the sigmoid raises dev accuracy from \CpAccYesNo{}\% to \CpAccSig{}\% and lowers the calibration error from \CpEceYesNo{} to \CpEceSig{}.
It also reduces label bias.
On the three datasets whose options are the same labels in every row (ContractNLI, MuSiQue and ProofWriter), we compare how often the model predicts each label with how often that label is correct.
The total variation (TV) between the two is the share of predictions that would have to change label for them to match, so 0 means no bias towards any label.
It is \CpLabelYes{} for the softmax over ``yes'', \CpLabelYesNo{} for the softmax over ``yes'' minus ``no'' and \CpLabelSig{} for the sigmoid; for example, on MuSiQue the softmax models predict ``Yes'' far more often than it is correct (\cref{sec:labels}).
TV measures bias, not accuracy: a model could predict each label exactly as often as it is correct and still assign it to the wrong rows.
The gold option's sigmoid averages \CpSigGoldSig{}; for small sigmoid values $\sigma(s)\approx e^{s}$, so the normalised sigmoid then approximates a softmax.

The square-root softplus has the lowest maximum gradient norm (\CpGradSqrt{}) but changes its answer most often under reordering (\CpShufSqrt{}\%).

\begin{table}[t]
  \centering\small
  \caption{Sensitivity to option order and removal, on the dev set. Each row is scored in four random option orders; an
  answer change under reordering means the chosen option is not the same in all four. One wrong option, never the
  chosen one, is removed at random from rows with at least three options, and the answer is compared over the remaining
  options. Dev accuracy is on the first random order, so it differs slightly from \cref{tab:headline}, which keeps
  the original order. TV: total variation between the predicted and the gold frequencies, over the three datasets with
  a fixed label set (label) and over option positions (position). Calibration error: expected calibration error of the
  top option's probability. $|s_k|$: size of the raw option scores. Each metric is averaged over datasets, then over
  training runs. Bold: best in the row, higher for accuracy and lower for answer changes, TV and calibration error; the
  gradient norm, the largest over all training steps and so a single batch's value, and $|s_k|$ are not ranked.}
  \label{tab:coupling}
\begin{tabular}{@{}lcccc@{}}
\toprule
& \begin{tabular}[b]{@{}c@{}}Softmax,\\``yes''\end{tabular} & \begin{tabular}[b]{@{}c@{}}Softmax,\\``yes'' $-$ ``no''\end{tabular} & \begin{tabular}[b]{@{}c@{}}Sqrt-softplus,\\``yes'' $-$ ``no''\end{tabular} & \begin{tabular}[b]{@{}c@{}}Sigmoid-norm,\\``yes'' $-$ ``no''\end{tabular} \\
\midrule
Dev accuracy (\%) & 71.4 & 71.5 & 71.1 & \textbf{72.9} \\
External accuracy (\%) & 57.8 & \textbf{59.6} & 59.3 & 59.5 \\
\midrule
Answer changes, options reordered (\%) & 15.6 & 14.6 & 16.6 & \textbf{13.7} \\
Answer changes, one wrong option removed (\%) & 5.9 & \textbf{4.8} & 5.4 & 4.9 \\
Largest gradient norm in training & 842 & 139 & 104 & 190 \\
\midrule
Label frequencies vs.\ gold (TV) & 0.167 & 0.213 & 0.154 & \textbf{0.139} \\
Position frequencies vs.\ gold (TV) & 0.050 & 0.045 & 0.049 & \textbf{0.041} \\
Calibration error, top option & 0.038 & 0.027 & 0.032 & \textbf{0.023} \\
Mean $|s_k|$ & 60.5 & 3.0 & 12.3 & 7.4 \\
\bottomrule
\end{tabular}

\end{table}

\subsection{Time and memory}
\label{sec:cost}

A decoder must read the whole trie: every option's score includes the closing bracket and, from ten options on, the later digits.
For a fair comparison we time the decoders with a faster scorer that reads the input once and then all trie nodes in one more forward pass on the shared KV cache, under a mask that lets each node see the input and its own earlier tokens.
In fp32 its scores match those of the evaluation scorer within $10^{-4}$ (40 dev rows with ten or more options); in bf16, as timed, both choose the same option on 1{,}297 of the 1{,}300 dev rows.
The evaluation scorer copies the cache for each node and takes \MsCopyQwFull{}\,ms and up to \MemPeakCopyQwFull{}\,GiB on Qwen3-0.6B.

One row at a time, the encoder takes \MsUb{}\,ms per decision, against \MsQwFull{}\,ms for Qwen3-0.6B and \MsQwFourLora{}\,ms for Qwen3-4B (\cref{tab:memory}); reading the input alone takes the decoders \MsPreQwFull{} and \MsPreQwFourLora{}\,ms.
Peak memory tells a different story.
Qwen3-0.6B peaks at \MemPeakQwFull{}\,GiB, against \MemPeakUb{}\,GiB for the encoder with the fp32 weights it needs for its accuracy, or \MemPeakUbBf{}\,GiB with bf16 weights, which lose accuracy (\cref{sec:precision}).
At this scale the encoder is faster but has no peak-memory advantage; we measured peaks, not memory over time.

\begin{table}[t]
  \centering\small
  \caption{Time and peak GPU memory per decision, scoring the dev rows one at a time on one A100. The decoders
  use bf16 weights and read the input, then the whole trie of option numbers in one more pass on the shared KV cache.
  Memory is what PyTorch allocates, in GiB, weights included; rows $>$4{,}096 tok.: mean peak over rows longer than
  4{,}096 tokens; max: the largest peak over all rows.}
  \label{tab:memory}
\begin{tabular}{@{}lcccc@{}}
\toprule
Model & ms per decision & Weights & Peak, rows $>$4{,}096 tok. & Peak, max \\
\midrule
UltraBERT-380M, fp32 weights & 29 & 1.6 & 2.6 & 2.8 \\
UltraBERT-380M, bf16 weights (lower accuracy) & 26 & 0.8 & 1.5 & 1.9 \\
Qwen3-0.6B & 68 & 1.1 & 2.0 & 2.2 \\
Qwen3-4B & 124 & 7.6 & 8.9 & 9.3 \\
\bottomrule
\end{tabular}

\end{table}

\section{Analysis}
\label{sec:analysis}

The numbers in this section are from the dev set, with one training run per setting, unless noted.

\subsection{One change at a time}

\Cref{tab:ablation} changes one part of the recipe at a time.
On these single runs, ``yes'' minus ``no'' adds 0.4 dev points and 2.0 external points, and the sigmoid a further 1.3 dev points, most of them on MuSiQue (64\% to 76\%).
Single runs on 100 rows per dataset are noisy: here Nimble falls from 55\% to 43\% with the sigmoid, while on the test set the sigmoid-normalised model is 3.7 points above the softmax over ``yes'' on Nimble.
The words ``correct'' and ``incorrect'' in place of ``yes'' and ``no'' did not help (70.6\% against 71.1\% for the single word, 70.1\% against 71.5\% for the difference).

\begin{table}[t]
  \centering\small
  \caption{Encoder variants on the dev rows, one training run each (accuracy, \%). The first row is the softmax with the ``yes'' logit,
  the setting used for the softmax results elsewhere. Bold: best in each column.}
  \label{tab:ablation}
\begin{tabular}{@{}lccccc@{}}
\toprule
Variant & Dev ALL & External & MuSiQue & ContractNLI & Nimble \\
\midrule
Softmax, ``yes'' & 71.1 & 57.1 & 57.0 & 75.0 & \textbf{55.0} \\
Softmax, ``yes'' $-$ ``no'' & 71.5 & 59.1 & 64.0 & 78.0 & 48.0 \\
Sigmoid-norm, ``yes'' $-$ ``no'' & \textbf{72.8} & \textbf{59.4} & \textbf{76.0} & \textbf{81.0} & 43.0 \\
Softmax, ``correct'' & 70.6 & 56.8 & 48.0 & 77.0 & 52.0 \\
Softmax, ``correct'' $-$ ``incorrect'' & 70.1 & 58.0 & 54.0 & 76.0 & 46.0 \\
\bottomrule
\end{tabular}

\end{table}

\subsection{Predicted labels}
\label{sec:labels}

The recipes predict different labels.
On MuSiQue, ``Yes'' is correct for 48\% of the dev rows; the sigmoid-normalised models predict it for 47 to 60\%, and the two softmax recipes for 64\% and 69\%.
On ProofWriter, ``Unknown'' is correct for a third of the rows; the softmax never predicts it and the sigmoid-normalised models predict it for 2 to 12\%, so it stays largely unlearned.

\subsection{Numerical precision}
\label{sec:precision}

The encoder's masked-token logits are around 200, where bf16 steps by 1 or 2 and options would tie or swap, so we compute the masked-token head, the scores and the normalisation in fp32.
Even then, bf16 weights changed MuSiQue accuracy by up to 8 points, presumably through rounding elsewhere in the network; we did not trace where.
We therefore keep fp32 weights under bf16 autocast.
Re-scoring a Qwen checkpoint in another precision moved its dev accuracy by at most 0.8 points.

\section{Predicting supporting evidence}
\label{sec:citation}

We add a small auxiliary head on the frozen encoder that highlights, in the same forward pass, the document sentences annotated as supporting the chosen option.
For each document token it reads the token's hidden state and that of the chosen option's mask, and outputs an evidence score between 0 and 1; a sentence takes the highest score among its tokens.
The score is not a calibrated probability, because positive tokens are up-weighted in training, and the head is not evaluated as an explanation of how the encoder decides.

\begin{figure}[t]
  \centering
  \begin{tikzpicture}[x=1in, y=1in,
      font=\sffamily\scriptsize,
      box/.style={draw=ink2, line width=0.4pt, rounded corners=2pt, inner sep=3.5pt, align=center, fill=black!2},
      key/.style={draw=encoder, line width=0.6pt, rounded corners=2pt, inner sep=3.5pt, align=center, fill=encoder!8},
      arr/.style={-{Stealth[length=4pt]}, line width=0.6pt, color=ink2},
      lane/.style={font=\sffamily\bfseries\scriptsize, text=ink2, anchor=west},
      sq/.style={draw=ink2, line width=0.3pt, minimum size=0.085in, inner sep=0pt}]
    \node[lane] at (0, 0.2) {1  Propose evidence sets};
    \node[box, text width=1.0in] (row) at (0.55, -0.35) {\textbf{Training row}\\document, question,\\options, gold answer};
    \node[box, text width=1.0in, right=0.2in of row] (sent) {\textbf{Sentences}\\the document split\\into sentences};
    \node[box, text width=1.8in, right=0.2in of sent] (ask) {\textbf{Ask the labelling model}\\Qwen3.6-27B with thinking, 10 samples\\(temperature 1.2). Prompt: numbered sentences,\\question, gold answer; ``cite the smallest set\\that supports it, without which\\it is unsupported''};
    \node[box, text width=1.08in, right=0.2in of ask] (cand) {\textbf{Candidate sets}\\merge duplicates; drop\\sets over 384 tokens\\unless a single sentence};
    \draw[arr] (row) -- (sent); \draw[arr] (sent) -- (ask); \draw[arr] (ask) -- (cand);
    \node[lane] at (1.25, -1.05) {2  Score every set on the decision prompt};
    \node[box, text width=1.75in, anchor=north west] (abl) at (0.05, -1.25) {%
      \textbf{Ablate each set $S$}\\[2pt]
      \begin{tabular}{@{}l@{\hspace{3pt}}l@{}}
        \tikz[baseline=-0.5ex]{\foreach \i/\c in {1/ink2!55,2/ink2!55,3/ink2!55,4/ink2!55,5/ink2!55,6/ink2!55,7/ink2!55}{\node[sq, fill=\c] at (\i*0.1, 0) {};}} & full context \\[1pt]
        \tikz[baseline=-0.5ex]{\foreach \i/\c in {1/ink2!55,2/white,3/ink2!55,4/ink2!55,5/white,6/ink2!55,7/ink2!55}{\node[sq, fill=\c] at (\i*0.1, 0) {};}} & without $S$ \\[1pt]
        \tikz[baseline=-0.5ex]{\foreach \i/\c in {1/white,2/ink2!55,3/white,4/white,5/ink2!55,6/white,7/white}{\node[sq, fill=\c] at (\i*0.1, 0) {};}} & only $S$
      \end{tabular}};
    \node[box, text width=1.35in, right=0.2in of abl] (q) {\textbf{Probability of the gold answer}\\$q(\mathrm{gold}\mid\cdot)$: softmax over the\\log-probabilities of the options};
    \node[key, text width=1.3in, right=0.2in of q] (rew) {\textbf{Reward (from SelfCite)}\\$\log q(\mathrm{gold}\mid\text{only }S)$\\$-\log q(\mathrm{gold}\mid\text{without }S)$\\sufficiency + necessity};
    \node[key, text width=1.0in, right=0.2in of rew] (out) {\textbf{Evidence label}\\highest reward\\$\to$ the row's spans};
    \draw[arr] (cand.south) -- ++(0, -0.17) -| (abl.north);
    \draw[arr] (abl) -- (q); \draw[arr] (q) -- (rew); \draw[arr] (rew) -- (out);
  \end{tikzpicture}
  \caption{How evidence labels are generated for rows without human evidence. The labelling model, Qwen3.6-27B, proposes
  up to ten sets of sentences as evidence for the \emph{gold} answer. The same model then answers the decision prompt
  three ways for each set $S$: with the full document, without $S$, and with only $S$: a good set is enough to keep the answer (sufficiency) and its removal loses the answer (necessity).
  The set with the highest reward becomes the label, which trains the evidence head. Unlike SelfCite, the answer scored is
  the gold answer and not the model's own response.}
  \label{fig:spans}
\end{figure}
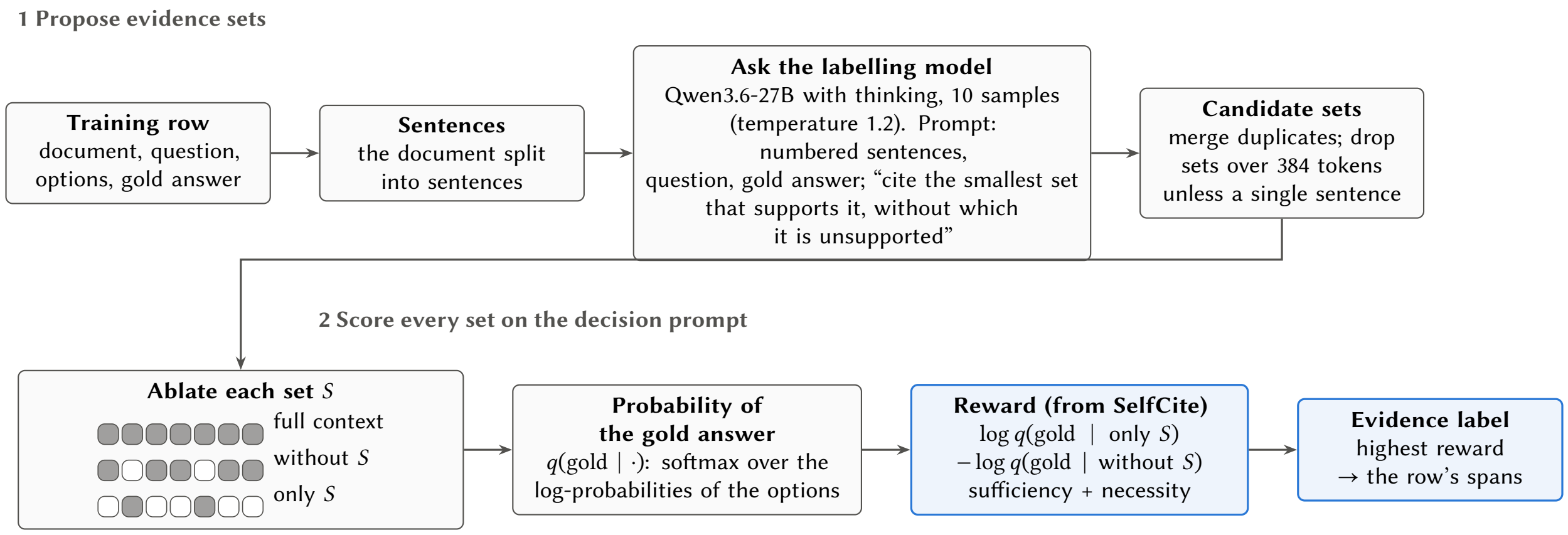

The head is trained on human-marked evidence where it exists, in ContractNLI, MuSiQue, Nimble and ProofWriter.
For the other rows, a labelling model proposes sets of sentences for the gold answer, and we keep the set with the best reward adapted from SelfCite \citep{selfcite} (\cref{fig:spans}, \cref{app:evidence}).

\paragraph{Comparison.}
We compare with the encoder's average attention, AT2 \citep{at2} and ContextCite \citep{contextcite} on the \EvRows{} test rows with gold evidence; the word-highlighting threshold is chosen on \EvTuneRows{} dev rows.
The head is trained on evidence labels, while AT2 and ContextCite are fitted to the model's response to ablations.

We report sentence average precision (AP) against the gold sentences, how often the top sentence is gold, and the time per row.
For the baselines we also report LDS \citep{contextcite}, the rank correlation between predicted and actual changes in the answer's probability when sentences are removed; we did not evaluate it for the heads.

AT2 and ContextCite ran on the softmax encoder, so the like-for-like comparison is the head on that encoder; the head on the sigmoid-normalised encoder also changes the encoder.
The heads see the option the encoder chose, and a wrong choice is scored against the gold option's evidence, so \cref{tab:citation} measures the decision and the evidence together.
The times are on these rows, so they cannot be subtracted from the decision time of \cref{sec:cost}.

\begin{table}[t]
  \centering\small
  \setlength{\tabcolsep}{4pt}
  \caption{Sentence evidence on the test rows with gold evidence. Sentence AP: average precision
  against the gold sentences, over all rows and per dataset (CNLI: ContractNLI, PW: ProofWriter). Top-1: share of rows
  where the top sentence is gold. LDS: how closely the scores follow the model's response to removed sentences; not evaluated for the heads. The first
  three methods and the first head use the encoder trained with the softmax. Each head is one training run. Bold: best in each column (lowest for time).}
  \label{tab:citation}
\begin{tabular}{@{}lcccccccc@{}}
\toprule
& \multicolumn{5}{c}{Sentence AP} & & & \\
\cmidrule(lr){2-6}
Method & All & CNLI & MuSiQue & Nimble & PW & Top-1 (\%) & LDS & ms per row \\
\midrule
Average attention & 0.25 & 0.07 & 0.20 & 0.34 & 0.38 & 14 & 0.09 & 69 \\
AT2 & 0.33 & 0.30 & 0.28 & 0.30 & 0.42 & 23 & 0.15 & 71 \\
ContextCite & 0.41 & 0.52 & 0.33 & 0.33 & 0.44 & 44 & \textbf{0.64} & 1{,}860 \\
\midrule
Evidence head, softmax encoder & 0.55 & 0.28 & 0.32 & \textbf{0.96} & 0.67 & 54 & -- & \textbf{68} \\
Evidence head, sigmoid-norm encoder & \textbf{0.68} & \textbf{0.55} & \textbf{0.52} & \textbf{0.96} & \textbf{0.68} & \textbf{74} & -- & 72 \\
\bottomrule
\end{tabular}

\end{table}

\paragraph{Results.}
On the softmax encoder, the head has a higher sentence AP than ContextCite and AT2 (\EvSentAPList{} against \ContextCiteSentAP{} and \AtTwoSentAP{}), in \EvMsList{}\,ms per row against \ContextCiteMs{}\,ms for ContextCite.
The lead comes from Nimble and ProofWriter; ContextCite is better on ContractNLI and level on MuSiQue.
The head on the sigmoid-normalised encoder reaches \EvSentAPSig{} and is level with or above ContextCite on those two datasets.

The words this head marks have a precision of \EvWordPSig{} and a recall of \EvWordRSig{}, a rough guide rather than an exact span.
Most of the accuracy comes from the human-marked labels: on the softmax encoder, a head trained on them alone reaches \EvSentAPGold{}, and the generated labels raise it to \EvSentAPList{} (\cref{app:evidence}).
Of the methods with an LDS, ContextCite follows the model most closely (\ContextCiteLds{}).

\section{Related work}
\label{sec:related}

\paragraph{Jev-style decision models.}
Most community decision models compared in LLM-as-Jev \citep{li2026llmasjev} read the logits of an option letter or label from a decoder, frozen or tuned with LoRA; LLM-as-Jev studies this readout and when fine-tuning helps.
AnyJev reads a typed decision from one prefill and corrects the label prior and option-order effects without training \citep{zhang2026anyjev}.
The Contrastive Language Model \citep{kwok2026clm} embeds the state and each candidate separately with a frozen Qwen3-8B and scores them by cosine similarity, so candidates can be cached but never attend to the context.
All of these start from a decoder; we start from an encoder that reads the context, the question and all options together.

\paragraph{Encoders that score options.}
Prompt-based classifiers use the masked-token head of an encoder with a verbalizer for a fixed set of labels \citep{schick2021pet, gao2021lmbff}.
Two encoder decision models are listed on public Jev leaderboards \citep{jevbench}: Laya \citep{laya} and Von \citep{von}, both built on ModernBERT-large with one mask per option.
Both add new layers, described in model cards: Laya trains its head with reinforcement learning on proper scoring rules, and Von uses a listwise softmax loss with a Brier term.
We reuse the pretrained masked-token head, compare four recipes against decoders trained on the same data, and report a held-out test split.

\paragraph{Option letters, position and label bias.}
Reading the letter of the chosen option requires a model to bind the letter to the option text \citep{robinson2023mcsb}.
Models favour certain option labels, and their accuracy changes when the options are reordered \citep{zheng2024notrobust, pezeshkpour2024order}.
Corrections are usually applied after training, by calibrating with a content-free input \citep{zhao2021calibrate} or by dividing out a label prior and averaging over rotations of the options \citep{zhang2026anyjev}.
We shuffle the options during training and measure how the recipes differ in the labels they predict.

\paragraph{Encoders and decoders.}
Controlled studies pretrain encoders and decoders with the same recipe, and find that encoders do well on classification and retrieval \citep{weller2025ettin} and that masked language modelling suits tasks that need text representations \citep{gisserotboukhlef2025mlm}.
A much larger decoder can still beat an encoder on classification \citep{zhao2024classification}.
Decoder designs such as multi-query and grouped-query attention target long generation \citep{mqa, gqa}, and \citet{prefillonly} study what they cost when nothing is generated (\cref{sec:intro}).
Modern encoders keep long inputs and fast inference \citep{modernbert, ultrabert}.

\paragraph{Attribution and evidence.}
ContextCite learns a surrogate model from many ablations of the context \citep{contextcite}.
AT2 learns to weight attention heads, and needs a decoder pass \citep{at2}.
SelfCite fine-tunes a model to cite, using the change in the answer probability when the cited text is removed or kept alone as the reward \citep{selfcite}.
Rationale benchmarks such as ERASER compare selected evidence with human rationales \citep{deyoung2020eraser}.
Our evidence head is trained on human-marked evidence and SelfCite-style labels, and gives the evidence in the decision's forward pass.

\section{Conclusion}
\label{sec:conclusion}

We fine-tuned an encoder, UltraBERT-380M, as a decision model with one mask per option, and compared four training recipes.
On the same training examples, every recipe is more accurate than Qwen3-0.6B trained with LLM-as-Jev (\TestUbList{}\% to \TestUbSig{}\% against \TestQwFull{}\% on the test set).
The encoder is faster (\MsUb{} against \MsQwFull{}\,ms per decision) but uses more peak memory (\MemPeakUb{} against \MemPeakQwFull{}\,GiB).

Among the recipes, scoring by ``yes'' minus ``no'' gives the larger test gain and all of the external gain.
Sigmoid normalisation adds \DiffUbSigUbYesNoAbs{} test points, mostly on MuSiQue, and brings label frequencies closer to the gold ones; the square-root softplus does not, and we cannot say which property of the sigmoid is responsible.
An auxiliary head on the frozen encoder predicts annotated evidence in the same pass, and on the softmax encoder it ranks the annotated sentences higher than AT2 and ContextCite, except on ContractNLI.

The results have limits.
They cover one encoder, one decoder family and one English training mixture, below one billion parameters; a 4B decoder is still more accurate except on the intent datasets.
We did not evaluate whether the evidence head reflects what the model used, and we measured only peak memory.

Two directions follow: larger encoders, to see whether the advantage holds above one billion parameters, and training the decision and the evidence together, so that the evidence head can shape the encoder.

\bibliographystyle{plainnat}
\bibliography{refs}

\appendix
\section{Implementation details}
\label{app:impl}

\Cref{tab:hparams} lists the training settings.
Both models train for one epoch, which is 406 steps at a batch of 32.

\begin{table}[h]
  \centering\small
  \caption{Training settings. Learning rates are in \cref{sec:Implementation details}.}
  \label{tab:hparams}
  \begin{tabular}{@{}lll@{}}
    \toprule
    & UltraBERT-380M & Qwen3-0.6B (full) \\
    \midrule
    Optimiser & AdamW, $\beta=(0.9, 0.98)$, $\epsilon=10^{-6}$ & AdamW \\
    Weight decay & 0.1 & 0.01 \\
    Schedule & 5\% warmup, linear decay to 0 & 100 warmup steps, linear decay to 10\% \\
    Gradient clipping & 1.0 & 1.0 \\
    Precision & fp32 weights, bf16 autocast & fp32 weights, bf16 autocast \\
    Attention & PyTorch SDPA & PyTorch SDPA \\
    Gradient checkpointing & yes & yes \\
    Token limit & 32{,}768 & 16{,}384 \\
    \bottomrule
  \end{tabular}
\end{table}

The encoder settings are those of the UltraBERT model card, and we change only the learning rate.
No training row exceeds either token limit, because the data are cut at \MaxTokens{} tokens (\cref{sec:data}).

\paragraph{Batched throughput.}
These numbers are specific to our implementations, not a comparison of the fastest paths.
When the dev rows are scored in batches, the encoder processes \RpsUb{} rows per second, against \RpsQwFull{} for Qwen3-0.6B and \RpsQwFourLora{} for Qwen3-4B.
The encoder uses packed variable-length FlashAttention-2 \citep{flashattention2} and the decoders use length-sorted padded batches with the cache-copying trie scorer, so part of this gap comes from the padding and the copies.

\section{Evidence head}
\label{app:evidence}

We train one head on each fine-tuned encoder (sigmoid-normalised and softmax, both at a learning rate of $1.5\cdot10^{-5}$); the softmax one matches the encoder used by AT2 and ContextCite.

\paragraph{Model.}
For document token $t$ and option $k$, the head reads $[h_t;\,h_k;\,h_t\odot h_k]$, the final hidden states of the token and of option $k$'s mask.
A two-layer perceptron (512 hidden units, GELU, dropout 0.1) maps it to one logit, whose sigmoid is the token's evidence score for option $k$.

\paragraph{Labels.}
A token is positive if it overlaps an evidence span.
We use gold evidence where it exists (Nimble, ContractNLI, MuSiQue, ProofWriter) and generated spans elsewhere (\cref{fig:spans}): ANLI, QuALITY, LogiQA~2.0, CosmosQA and the remaining rows of ContractNLI, MuSiQue and ProofWriter.
There are 7{,}778 labelled training rows (7{,}390 for training, 388 for validation), with 5.2\% positive tokens.

\paragraph{Generated spans.}
Qwen3.6-27B with thinking gets the numbered sentences, the question and the gold answer, and is asked for the smallest set of sentences that supports the answer and without which it is unsupported.
We sample ten replies (temperature 1.2, top-$p$ 0.9), merge identical sets, and drop sets over 384 tokens unless they are one sentence.

\paragraph{Reward.}
The same model scores the decision prompt with each set $S$ removed and with only $S$ kept; $q(\mathrm{gold}\mid\cdot)$ is the softmax over the options' summed token log-probabilities.
The label is the set with the highest reward $\log q(\mathrm{gold}\mid\text{only }S)-\log q(\mathrm{gold}\mid\text{without }S)$, which is sufficiency plus necessity, with the full-context term cancelling.
The best set is always kept, so a row whose candidates are all poor still gets a label, and nothing beyond the token cap makes the set the smallest sufficient one.
Unlike SelfCite, we score the gold answer, so the reward is informative only where the labelling model believes it.
We did not check the generated labels against human evidence directly; the last paragraph of this appendix is an indirect check.

\paragraph{Training.}
Token-level binary cross-entropy, with positives weighted by the negative-to-positive ratio (18.1), on the frozen encoder.
The head is conditioned on the gold option in training and on the chosen option at test time.
AdamW (learning rate $10^{-3}$, weight decay 0.01), 16 rows per step, three epochs; we keep the epoch with the best validation token average precision.

\paragraph{Evaluation.}
All methods are scored on the test rows with gold evidence and at least two sentences.
A word is highlighted when its evidence score reaches a threshold tuned for word F1 on the \EvTuneRows{} dev rows with gold evidence: \EvThrSig{} for the sigmoid-normalised encoder and \EvThrList{} for the softmax encoder.
On the sigmoid-normalised encoder, the head's token average precision is \EvTokAPSig{} (\EvChanceSig{} for a random ranking), and \EvPTenSig{} of its ten highest tokens are gold.
On the softmax encoder, adding the generated labels to the gold ones raises sentence average precision from \EvSentAPGold{} to \EvSentAPList{} and token average precision from \EvTokAPGold{} to \EvTokAPList{}.

\section{Additional results}
\label{app:extra}

\Cref{tab:perdataset} gives the test accuracy for each dataset.

\begin{table}[h]
  \centering\small
  \setlength{\tabcolsep}{4pt}
  \caption{Test accuracy (\%) per dataset, averaged over training runs. UB is UltraBERT-380M: UB yes is the softmax over each option's ``yes'' logit; UB yes$-$no is the softmax over ``yes'' minus ``no''; UB sqrt-sp and UB sig-norm normalise ``yes'' minus ``no'' with the square-root softplus and the sigmoid, each divided by the sum over options (\cref{sec:loss}). Bold: best below 1B parameters.}
  \label{tab:perdataset}
\begin{tabular}{@{}lrccccccc@{}}
\toprule
Dataset & Rows & UB yes & UB yes$-$no & UB sqrt-sp & UB sig-norm & Qwen full & Qwen LoRA & Qwen-4B LoRA \\
\midrule
CLINC150 & 400 & \textbf{91.8} & 91.5 & 91.7 & \textbf{91.8} & 86.3 & 79.8 & 89.8 \\
MASSIVE & 400 & 83.7 & 84.5 & 83.8 & \textbf{84.9} & 76.9 & 68.8 & 83.5 \\
Bitext & 400 & \textbf{99.3} & \textbf{99.3} & \textbf{99.3} & \textbf{99.3} & 96.0 & 94.8 & 97.2 \\
\midrule
ReClor & 400 & 58.9 & 60.7 & \textbf{61.8} & 59.9 & 57.7 & 51.2 & 72.2 \\
LogiQA 2.0 & 400 & 52.9 & 54.8 & 55.4 & 54.8 & \textbf{57.0} & 56.0 & 73.5 \\
CosmosQA & 400 & 74.8 & 74.2 & 75.1 & \textbf{75.2} & 59.9 & 55.0 & 76.0 \\
PIQA & 400 & 74.8 & \textbf{77.5} & 75.9 & 75.1 & 70.2 & 68.8 & 83.8 \\
ANLI & 400 & 73.1 & 72.6 & \textbf{73.9} & 73.6 & 63.7 & 61.0 & 74.0 \\
ProofWriter & 400 & 60.7 & 61.5 & 61.4 & 61.6 & \textbf{63.3} & 63.0 & 70.5 \\
\midrule
ContractNLI & 393 & 73.8 & 77.4 & 74.7 & \textbf{78.6} & 76.1 & 65.6 & 79.1 \\
MuSiQue & 400 & 58.9 & 63.0 & 62.2 & \textbf{69.8} & 62.9 & 54.8 & 71.5 \\
QuALITY & 398 & 56.9 & \textbf{62.1} & 60.8 & 61.3 & 51.5 & 50.0 & 68.8 \\
Nimble & 224 & 47.0 & 47.8 & 49.1 & 50.7 & \textbf{57.3} & 50.9 & 64.3 \\
\bottomrule
\end{tabular}

\end{table}

\section{Examples}
\label{app:examples}

Each example shows the sigmoid-normalised encoder's decision and its evidence head's token scores, on the first test row of the evidence study with at most 1{,}400 tokens in each dataset (MuSiQue has none).
Tokens at or above the dev-tuned threshold of \EvThrSig{} are highlighted, darker for a higher score, and gold evidence is underlined.
The encoder is correct on all three rows, which were picked by this rule, not by outcome.

\paragraph{ContractNLI.} 1{,}338 tokens.
The model chooses option 1, which is \textbf{correct}. Question: Statement: "Receiving Party shall not reverse engineer any objects which embody Disclosing Party's Confidential Information." Does the contract entail the statement, contradict it, or not mention it?

\begin{tabular}{@{}rrp{0.7\linewidth}l@{}}
\toprule
$k$ & Probability & Option & \\
\midrule
1 & 85\% & The contract entails the statement. & gold, chosen \\
2 & 0\% & The contract contradicts the statement. &  \\
3 & 15\% & The contract does not mention it. &  \\
\bottomrule
\end{tabular}

\begingroup\scriptsize\setlength{\parindent}{0pt}\setlength{\parskip}{0.3em}\raggedright
\textsf{\textbf{Contract}}\quad MUTUAL NONDISCLOSURE AGREEMENT Effective Date: 12/10/98 This Agreement governs the disclosure of information by and between Yahoo! Inc., a California corporation, and Restrac, Inc. ("Participant"). 1. The "Confidential Information" is that confidential{\fboxsep=0pt\colorbox[HTML]{B2CEF0}{,}} proprietary, and {\fboxsep=0pt\colorbox[HTML]{BAD3F2}{trade}} secret information being disclosed by the disclosing party described as (please be specific): (a) {\fboxsep=0pt\colorbox[HTML]{C5DBF4}{Yahoo}} Confidential Information (owned by Yahoo and any of its affiliates): Resume and Employment Related. (b) Participant Confidential Information: Resume and Employment Related. 2. Except as set forth in this Section 2, all Confidential Information shall be in {\fboxsep=0pt\colorbox[HTML]{B5D0F1}{tangible}} form and shall be marked as Confidential or {\fboxsep=0pt\colorbox[HTML]{A8C8EE}{proprietary}} information of the disclosing party. If the Confidential Information is disclosed orally or visually, it shall be identified {\fboxsep=0pt\colorbox[HTML]{ADCBEF}{as}} such at the time of disclosure and confirmed in a writing to the recipient within thirty (30) days of such disclosure. 3. Each of the parties agrees that it will {\fboxsep=0pt\colorbox[HTML]{B5D0F1}{not}} make use of, disseminate, or in any way disclose any Confidential Information of the other party to any person, firm or business, except to the extent necessary for negotiations, discussions and consultations with personnel or authorized representatives of the other party and any purpose the other party may hereafter authorize in writing. Each of the parties agrees that it shall disclose Confidential Information of the other party only to those of its employees who need to know such information and who have previously agreed, either as a condition to employment or in order to obtain the Confidential Information, to be bound by terms and conditions substantially similar to those of this Agreement. 4. There shall be no liability for disclosure or use of Confidential Information which is (a) in the public domain through no fault of the receiving party (b) rightfully received from a third party without any obligation of confidentiality, (c) rightfully known to the receiving party without any limitation on use or disclosure prior to its receipt from the disclosing party, (d) independently developed by the receiving party, (e) generally made available to third parties without any restriction on disclosure, or (f) communicated in response to a valid order by a court or other governmental body, as otherwise required by law, or as necessary to establish the rights of either party under this Agreement (provided that the party so disclosing has provided the other party with a reasonable opportunity to seek protective legal treatment for such Confidential Information). 5. "Residual Information" shall mean any Confidential Information of the disclosing party which may be retained in intangible form in the minds of those individuals of the receiving party who have had proper access to such Confidential Information. Notwithstanding anything else in this Agreement, the receiving party shall be free to use any Residual Information for any purpose whatsoever, including, without limitation, the development of its own products, provided that such party shall not be entitled to disclose Residual Information to any third parties unless such disclosure is in the course of, or as part of, any disclosure of its own products or their development. 6. Each of the parties agrees that it shall treat all Confidential Information of the other party with the same degree of care as it accords to its own Confidential Information and each of the parties represents that it exercises reasonable care to protect its own Confidential Information. \underline{7}\underline{.} \underline{Each} \underline{of} \underline{the} \underline{parties} \underline{agrees} \underline{that} \underline{it} \underline{will} \underline{not} \underline{modify}{\fboxsep=0pt\colorbox[HTML]{BFD6F3}{\underline{,}}} {\fboxsep=0pt\colorbox[HTML]{BED5F2}{\underline{reverse}}} {\fboxsep=0pt\colorbox[HTML]{A8C8EE}{\underline{engineer}}}\underline{,} \underline{de}{\fboxsep=0pt\colorbox[HTML]{9FC2ED}{\underline{compile}}}\underline{,} \underline{create} \underline{other} \underline{works} \underline{from}\underline{,} \underline{or} \underline{dis}{\fboxsep=0pt\colorbox[HTML]{A8C8EE}{\underline{assemble}}} {\fboxsep=0pt\colorbox[HTML]{A3C5ED}{\underline{any}}} {\fboxsep=0pt\colorbox[HTML]{A2C4ED}{\underline{software}}} {\fboxsep=0pt\colorbox[HTML]{A6C7EE}{\underline{programs}}} \underline{contained} {\fboxsep=0pt\colorbox[HTML]{ACCBEF}{\underline{in}}} \underline{the} \underline{Conf}\underline{idential} \underline{Information} \underline{of} \underline{the} \underline{other} \underline{party} \underline{unless} \underline{otherwise} \underline{specified} \underline{in} \underline{writing} \underline{by} \underline{the} \underline{disclosing} \underline{party}\underline{.} 8. All materials (including, without limitation, documents, drawings, models, apparatus, sketches, designs and lists) furnished to one party by the other, and which are designated in writing to be the property of such party, shall remain the property of such party and shall be returned to it promptly at its request, together with any copies thereof. 9. This Agreement shall govern all communications between the parties that are made during the period from the effective date of this Agreement to the date on which either party receives from the other written notice that subsequent communications shall not be so governed, provided, however that each party's obligations under Sections 2 and 3 with respect to Confidential Information of the other party which it has previously received shall continue unless and until such Confidential Information falls within Sections 4 or 5. 10. Neither party shall communicate any information to the other in violation of the proprietary rights of any third party. Neither party acquires any licenses under any intellectual property rights of the other party under this Agreement. This Agreement shall be governed in all respects by the laws of the United States of America and by the laws of the State of California as such laws are applied to agreements entered into and to be performed entirely within California between California residents. This Agreement may only be changed by mutual agreement of authorized representatives of the parties in writing. All notices or reports permitted or required under this Agreement shall be in writing and shall be by personal delivery, telegram, telex, telecopier, facsimile transmission or by certified or registered mail, return receipt requested, and shall be deemed given upon personal delivery, five (5) days after deposit in the mail, or upon acknowledgment of receipt of electronic transmission. Notices shall be sent to the addresses set forth at the end of this Agreement or such other address as either party may specify in writing. IN WITNESS WHEREOF, the parties have executed this Agreement in duplicate as of the date first written above. <TABLE> <S> <C> YAHOO! INC. -------------------------------------------- By: /s/ Matt Rowlen By: /s/ M. J. Fahey ------------------------------------- --------------------------------------- Printed Name: Matt Rowlen Printed Name: M. J. Fahey --------------------------- ----------------------------- Title: Manager of Business Development Title: President -------------------------------- ------------------------------------ Address: Yahoo! Address: Lexington, MA -------------------------------- ---------------------------------- -------------------------------- ---------------------------------- -------------------------------- ---------------------------------- </TABLE> 

\endgroup

\paragraph{Nimble.} 558 tokens.
The model chooses option 3, which is \textbf{correct}. Question: Apply this rubric: HIGH---route is correct, custody is intact, and the claimant supplies both a unique identifier and one undisclosed feature without a material conflict; release is permitted. MEDIUM---route is correct and the claim is plausible, but either required identifying element is missing or custody has a gap; hold for more evidence. LOW---route is wrong or a material item, date, location, or identifier conflicts; do not release. Weather, staff clothing, and similar incidental observations are distractors, not material conflicts.

\begin{tabular}{@{}rrp{0.7\linewidth}l@{}}
\toprule
$k$ & Probability & Option & \\
\midrule
1 & 6\% & Do not release: LOW confidence because the route is incorrect or material facts conflict with the intake record. &  \\
2 & 35\% & Send to the Transit Depot Vault but hold the backpack: MEDIUM confidence because an identifying requirement is missing or custody is incomplete. &  \\
3 & 60\% & Send to the Transit Depot Vault and release to Mara: HIGH confidence because custody is intact and both required identifying elements match without material conflict. & gold, chosen \\
\bottomrule
\end{tabular}

\begingroup\scriptsize\setlength{\parindent}{0pt}\setlength{\parskip}{0.3em}\raggedright
\textsf{\textbf{Context}}\quad {\fboxsep=0pt\colorbox[HTML]{96BDEB}{A}} {\fboxsep=0pt\colorbox[HTML]{7DACE6}{navy}} {\fboxsep=0pt\colorbox[HTML]{81AFE7}{canvas}} {\fboxsep=0pt\colorbox[HTML]{71A5E4}{backpack}} {\fboxsep=0pt\colorbox[HTML]{87B3E8}{was}} {\fboxsep=0pt\colorbox[HTML]{81AFE7}{found}} {\fboxsep=0pt\colorbox[HTML]{6DA2E3}{on}} {\fboxsep=0pt\colorbox[HTML]{9EC2EC}{Route}} {\fboxsep=0pt\colorbox[HTML]{84B1E7}{1}}{\fboxsep=0pt\colorbox[HTML]{7AAAE5}{8}} bus {\fboxsep=0pt\colorbox[HTML]{7CACE6}{6}}{\fboxsep=0pt\colorbox[HTML]{659DE1}{0}}{\fboxsep=0pt\colorbox[HTML]{689FE2}{4}} {\fboxsep=0pt\colorbox[HTML]{669EE2}{at}} {\fboxsep=0pt\colorbox[HTML]{69A0E2}{1}}{\fboxsep=0pt\colorbox[HTML]{679EE2}{8}}{\fboxsep=0pt\colorbox[HTML]{6CA2E3}{:}}{\fboxsep=0pt\colorbox[HTML]{74A7E4}{2}}{\fboxsep=0pt\colorbox[HTML]{679EE2}{0}} {\fboxsep=0pt\colorbox[HTML]{639CE1}{on}} {\fboxsep=0pt\colorbox[HTML]{669EE1}{1}}{\fboxsep=0pt\colorbox[HTML]{69A0E2}{2}} {\fboxsep=0pt\colorbox[HTML]{669EE2}{May}}{\fboxsep=0pt\colorbox[HTML]{77A8E5}{.}} {\fboxsep=0pt\colorbox[HTML]{94BBEA}{Transit}} {\fboxsep=0pt\colorbox[HTML]{A5C6EE}{operator}} I{\fboxsep=0pt\colorbox[HTML]{78AAE5}{man}}i {\fboxsep=0pt\colorbox[HTML]{85B2E8}{Cole}} {\fboxsep=0pt\colorbox[HTML]{80AEE6}{sealed}} it {\fboxsep=0pt\colorbox[HTML]{639CE1}{as}} {\fboxsep=0pt\colorbox[HTML]{689FE2}{T}}{\fboxsep=0pt\colorbox[HTML]{6BA1E2}{-}}{\fboxsep=0pt\colorbox[HTML]{6BA1E3}{8}}{\fboxsep=0pt\colorbox[HTML]{639CE1}{8}}{\fboxsep=0pt\colorbox[HTML]{659DE1}{4}} {\fboxsep=0pt\colorbox[HTML]{73A6E4}{and}} {\fboxsep=0pt\colorbox[HTML]{72A6E4}{delivered}} {\fboxsep=0pt\colorbox[HTML]{71A5E4}{it}} to the {\fboxsep=0pt\colorbox[HTML]{AFCCF0}{civic}} hall receptionist{\fboxsep=0pt\colorbox[HTML]{89B4E8}{,}} who {\fboxsep=0pt\colorbox[HTML]{7CACE6}{transferred}} the unbroken seal {\fboxsep=0pt\colorbox[HTML]{A8C8EE}{to}} {\fboxsep=0pt\colorbox[HTML]{AAC9EF}{lost}}{\fboxsep=0pt\colorbox[HTML]{ADCBEF}{-}}{\fboxsep=0pt\colorbox[HTML]{C3D9F3}{property}} custodian {\fboxsep=0pt\colorbox[HTML]{85B2E8}{Ar}}{\fboxsep=0pt\colorbox[HTML]{77A9E5}{un}} {\fboxsep=0pt\colorbox[HTML]{639CE1}{V}}{\fboxsep=0pt\colorbox[HTML]{82B0E7}{ale}} without any gap in {\fboxsep=0pt\colorbox[HTML]{98BEEB}{custody}}{\fboxsep=0pt\colorbox[HTML]{8AB5E8}{.}}

\textsf{\textbf{Facts}}\quad {\fboxsep=0pt\colorbox[HTML]{68A0E2}{The}} {\fboxsep=0pt\colorbox[HTML]{70A5E4}{backpack}} {\fboxsep=0pt\colorbox[HTML]{639CE1}{sealed}} {\fboxsep=0pt\colorbox[HTML]{639CE1}{as}} {\fboxsep=0pt\colorbox[HTML]{629BE1}{T}}{\fboxsep=0pt\colorbox[HTML]{629CE1}{-}}{\fboxsep=0pt\colorbox[HTML]{70A4E3}{8}}{\fboxsep=0pt\colorbox[HTML]{669EE2}{8}}{\fboxsep=0pt\colorbox[HTML]{659DE1}{4}} {\fboxsep=0pt\colorbox[HTML]{83B0E7}{was}} {\fboxsep=0pt\colorbox[HTML]{80AEE7}{found}} {\fboxsep=0pt\colorbox[HTML]{85B2E7}{in}} {\fboxsep=0pt\colorbox[HTML]{679FE2}{the}} {\fboxsep=0pt\colorbox[HTML]{92BAEA}{transit}} {\fboxsep=0pt\colorbox[HTML]{A5C6EE}{vehicle}} {\fboxsep=0pt\colorbox[HTML]{74A7E4}{Route}} {\fboxsep=0pt\colorbox[HTML]{79AAE5}{1}}{\fboxsep=0pt\colorbox[HTML]{639CE1}{8}} {\fboxsep=0pt\colorbox[HTML]{8AB5E8}{bus}} {\fboxsep=0pt\colorbox[HTML]{6DA3E3}{6}}{\fboxsep=0pt\colorbox[HTML]{7AABE5}{0}}{\fboxsep=0pt\colorbox[HTML]{689FE2}{4}}{\fboxsep=0pt\colorbox[HTML]{86B2E8}{.}} The {\fboxsep=0pt\colorbox[HTML]{78AAE5}{custody}} {\fboxsep=0pt\colorbox[HTML]{89B4E8}{chain}} {\fboxsep=0pt\colorbox[HTML]{6DA2E3}{for}} {\fboxsep=0pt\colorbox[HTML]{75A8E4}{backpack}} {\fboxsep=0pt\colorbox[HTML]{639CE1}{T}}{\fboxsep=0pt\colorbox[HTML]{669EE2}{-}}{\fboxsep=0pt\colorbox[HTML]{75A8E4}{8}}{\fboxsep=0pt\colorbox[HTML]{79AAE5}{8}}{\fboxsep=0pt\colorbox[HTML]{689FE2}{4}} has no gap between I{\fboxsep=0pt\colorbox[HTML]{7FAEE6}{man}}{\fboxsep=0pt\colorbox[HTML]{9DC1EC}{i}} {\fboxsep=0pt\colorbox[HTML]{79AAE5}{Cole}} {\fboxsep=0pt\colorbox[HTML]{B2CEF0}{sealing}} it and {\fboxsep=0pt\colorbox[HTML]{7CACE6}{Ar}}{\fboxsep=0pt\colorbox[HTML]{70A5E4}{un}} {\fboxsep=0pt\colorbox[HTML]{72A5E4}{V}}{\fboxsep=0pt\colorbox[HTML]{87B3E8}{ale}} {\fboxsep=0pt\colorbox[HTML]{94BBEA}{receiving}} {\fboxsep=0pt\colorbox[HTML]{81AFE7}{it}} {\fboxsep=0pt\colorbox[HTML]{81AFE7}{from}} {\fboxsep=0pt\colorbox[HTML]{86B3E8}{the}} {\fboxsep=0pt\colorbox[HTML]{B4CFF1}{civic}} hall receptionist{\fboxsep=0pt\colorbox[HTML]{8AB5E9}{.}} The seal {\fboxsep=0pt\colorbox[HTML]{75A8E4}{on}} {\fboxsep=0pt\colorbox[HTML]{78A9E5}{backpack}} {\fboxsep=0pt\colorbox[HTML]{609BE0}{T}}{\fboxsep=0pt\colorbox[HTML]{6FA3E3}{-}}{\fboxsep=0pt\colorbox[HTML]{78AAE5}{8}}{\fboxsep=0pt\colorbox[HTML]{6AA1E2}{8}}{\fboxsep=0pt\colorbox[HTML]{679EE2}{4}} remained {\fboxsep=0pt\colorbox[HTML]{B0CDF0}{unbroken}} through {\fboxsep=0pt\colorbox[HTML]{7FAEE6}{Ar}}{\fboxsep=0pt\colorbox[HTML]{72A6E4}{un}} {\fboxsep=0pt\colorbox[HTML]{71A5E4}{V}}{\fboxsep=0pt\colorbox[HTML]{70A5E4}{ale}}{\fboxsep=0pt\colorbox[HTML]{6BA1E2}{'s}} receipt {\fboxsep=0pt\colorbox[HTML]{7AABE5}{of}} it{\fboxsep=0pt\colorbox[HTML]{86B2E8}{.}} {\fboxsep=0pt\colorbox[HTML]{679FE2}{The}} {\fboxsep=0pt\colorbox[HTML]{609AE0}{serial}} {\fboxsep=0pt\colorbox[HTML]{679EE2}{patch}} {\fboxsep=0pt\colorbox[HTML]{6BA1E3}{recorded}} {\fboxsep=0pt\colorbox[HTML]{669EE2}{for}} {\fboxsep=0pt\colorbox[HTML]{73A6E4}{backpack}} {\fboxsep=0pt\colorbox[HTML]{619BE1}{T}}{\fboxsep=0pt\colorbox[HTML]{679FE2}{-}}{\fboxsep=0pt\colorbox[HTML]{75A8E5}{8}}{\fboxsep=0pt\colorbox[HTML]{669EE2}{8}}{\fboxsep=0pt\colorbox[HTML]{659DE1}{4}} {\fboxsep=0pt\colorbox[HTML]{89B5E8}{is}} {\fboxsep=0pt\colorbox[HTML]{70A5E4}{a}} {\fboxsep=0pt\colorbox[HTML]{84B1E7}{unique}} {\fboxsep=0pt\colorbox[HTML]{83B0E7}{identifier}} {\fboxsep=0pt\colorbox[HTML]{6EA3E3}{for}} {\fboxsep=0pt\colorbox[HTML]{81AFE7}{that}} {\fboxsep=0pt\colorbox[HTML]{7EADE6}{backpack}}{\fboxsep=0pt\colorbox[HTML]{8FB8E9}{.}} {\fboxsep=0pt\colorbox[HTML]{619BE1}{\underline{The}}} {\fboxsep=0pt\colorbox[HTML]{619BE1}{\underline{serial}}}{\fboxsep=0pt\colorbox[HTML]{6BA1E3}{\underline{-}}}{\fboxsep=0pt\colorbox[HTML]{75A7E4}{\underline{patch}}} {\fboxsep=0pt\colorbox[HTML]{619BE1}{\underline{value}}} {\fboxsep=0pt\colorbox[HTML]{629BE1}{\underline{recorded}}} {\fboxsep=0pt\colorbox[HTML]{639CE1}{\underline{for}}} {\fboxsep=0pt\colorbox[HTML]{6EA3E3}{\underline{backpack}}} {\fboxsep=0pt\colorbox[HTML]{609AE0}{\underline{T}}}{\fboxsep=0pt\colorbox[HTML]{74A7E4}{\underline{-}}}{\fboxsep=0pt\colorbox[HTML]{76A8E5}{\underline{8}}}{\fboxsep=0pt\colorbox[HTML]{6BA1E2}{\underline{8}}}{\fboxsep=0pt\colorbox[HTML]{679FE2}{\underline{4}}} {\fboxsep=0pt\colorbox[HTML]{6AA1E2}{\underline{in}}} {\fboxsep=0pt\colorbox[HTML]{7BABE6}{\underline{the}}} {\fboxsep=0pt\colorbox[HTML]{649DE1}{\underline{intake}}} {\fboxsep=0pt\colorbox[HTML]{6FA4E3}{\underline{log}}} {\fboxsep=0pt\colorbox[HTML]{649CE1}{\underline{is}}} {\fboxsep=0pt\colorbox[HTML]{649DE1}{\underline{SP}}}{\fboxsep=0pt\colorbox[HTML]{72A5E4}{\underline{-}}}{\fboxsep=0pt\colorbox[HTML]{619BE1}{\underline{7}}}{\fboxsep=0pt\colorbox[HTML]{609AE0}{\underline{7}}}{\fboxsep=0pt\colorbox[HTML]{619BE1}{\underline{4}}}{\fboxsep=0pt\colorbox[HTML]{619BE1}{\underline{2}}}{\fboxsep=0pt\colorbox[HTML]{81AFE7}{\underline{.}}} {\fboxsep=0pt\colorbox[HTML]{78AAE5}{\underline{M}}}{\fboxsep=0pt\colorbox[HTML]{B1CEF0}{\underline{ara}}} {\fboxsep=0pt\colorbox[HTML]{639CE1}{\underline{V}}}{\fboxsep=0pt\colorbox[HTML]{77A9E5}{\underline{enn}}} {\fboxsep=0pt\colorbox[HTML]{6EA3E3}{\underline{supplied}}} {\fboxsep=0pt\colorbox[HTML]{629BE1}{\underline{the}}} {\fboxsep=0pt\colorbox[HTML]{609AE0}{\underline{serial}}}{\fboxsep=0pt\colorbox[HTML]{6FA4E3}{\underline{-}}}{\fboxsep=0pt\colorbox[HTML]{7BABE6}{\underline{patch}}} {\fboxsep=0pt\colorbox[HTML]{629BE1}{\underline{value}}} {\fboxsep=0pt\colorbox[HTML]{619BE1}{\underline{SP}}}{\fboxsep=0pt\colorbox[HTML]{7AABE5}{\underline{-}}}{\fboxsep=0pt\colorbox[HTML]{649DE1}{\underline{7}}}{\fboxsep=0pt\colorbox[HTML]{629CE1}{\underline{7}}}{\fboxsep=0pt\colorbox[HTML]{649DE1}{\underline{4}}}{\fboxsep=0pt\colorbox[HTML]{629BE1}{\underline{2}}} {\fboxsep=0pt\colorbox[HTML]{B4CFF1}{\underline{when}}} \underline{she} \underline{filed} \underline{her} \underline{claim}{\fboxsep=0pt\colorbox[HTML]{7EADE6}{\underline{.}}} {\fboxsep=0pt\colorbox[HTML]{7AABE5}{The}} {\fboxsep=0pt\colorbox[HTML]{A2C4ED}{hidden}}{\fboxsep=0pt\colorbox[HTML]{84B1E7}{-}}{\fboxsep=0pt\colorbox[HTML]{B3CFF0}{feature}} {\fboxsep=0pt\colorbox[HTML]{7DACE6}{detail}} {\fboxsep=0pt\colorbox[HTML]{6BA1E2}{recorded}} {\fboxsep=0pt\colorbox[HTML]{6AA0E2}{for}} {\fboxsep=0pt\colorbox[HTML]{84B1E7}{backpack}} {\fboxsep=0pt\colorbox[HTML]{619BE1}{T}}{\fboxsep=0pt\colorbox[HTML]{74A7E4}{-}}{\fboxsep=0pt\colorbox[HTML]{6DA2E3}{8}}{\fboxsep=0pt\colorbox[HTML]{69A0E2}{8}}{\fboxsep=0pt\colorbox[HTML]{6BA1E3}{4}}{\fboxsep=0pt\colorbox[HTML]{80AFE7}{,}} {\fboxsep=0pt\colorbox[HTML]{B4CFF1}{a}} {\fboxsep=0pt\colorbox[HTML]{81AFE7}{brass}} {\fboxsep=0pt\colorbox[HTML]{87B3E8}{key}} {\fboxsep=0pt\colorbox[HTML]{7BABE6}{inside}} {\fboxsep=0pt\colorbox[HTML]{80AFE7}{a}} {\fboxsep=0pt\colorbox[HTML]{A6C6EE}{torn}} inner{\fboxsep=0pt\colorbox[HTML]{7EADE6}{-}}lining {\fboxsep=0pt\colorbox[HTML]{B0CDF0}{pocket}}{\fboxsep=0pt\colorbox[HTML]{75A8E4}{,}} {\fboxsep=0pt\colorbox[HTML]{9CC0EC}{was}} not publicly disclosed before {\fboxsep=0pt\colorbox[HTML]{7FAEE6}{Mar}}{\fboxsep=0pt\colorbox[HTML]{7CACE6}{a}} {\fboxsep=0pt\colorbox[HTML]{86B2E8}{V}}{\fboxsep=0pt\colorbox[HTML]{6FA4E3}{enn}} {\fboxsep=0pt\colorbox[HTML]{81AFE7}{supplied}} it{\fboxsep=0pt\colorbox[HTML]{89B4E8}{.}} M{\fboxsep=0pt\colorbox[HTML]{84B1E7}{ara}} {\fboxsep=0pt\colorbox[HTML]{B5D0F1}{V}}{\fboxsep=0pt\colorbox[HTML]{A7C7EE}{enn}} {\fboxsep=0pt\colorbox[HTML]{81AFE7}{supplied}} {\fboxsep=0pt\colorbox[HTML]{90B8EA}{a}} matching {\fboxsep=0pt\colorbox[HTML]{72A6E4}{brass}}{\fboxsep=0pt\colorbox[HTML]{72A6E4}{-}}{\fboxsep=0pt\colorbox[HTML]{73A6E4}{key}} {\fboxsep=0pt\colorbox[HTML]{689FE2}{hidden}}{\fboxsep=0pt\colorbox[HTML]{6EA3E3}{-}}{\fboxsep=0pt\colorbox[HTML]{7AABE5}{feature}} {\fboxsep=0pt\colorbox[HTML]{6DA2E3}{detail}}{\fboxsep=0pt\colorbox[HTML]{76A8E5}{.}} All other material {\fboxsep=0pt\colorbox[HTML]{BDD5F2}{items}}, dates{\fboxsep=0pt\colorbox[HTML]{77A9E5}{,}} times, locations{\fboxsep=0pt\colorbox[HTML]{8CB6E9}{,}} and identifiers {\fboxsep=0pt\colorbox[HTML]{7DADE6}{Mar}}{\fboxsep=0pt\colorbox[HTML]{73A6E4}{a}} {\fboxsep=0pt\colorbox[HTML]{9FC2EC}{V}}{\fboxsep=0pt\colorbox[HTML]{88B4E8}{enn}} {\fboxsep=0pt\colorbox[HTML]{7FAEE6}{supplied}} agree with {\fboxsep=0pt\colorbox[HTML]{86B3E8}{the}} {\fboxsep=0pt\colorbox[HTML]{629BE1}{intake}} {\fboxsep=0pt\colorbox[HTML]{7AAAE5}{record}}{\fboxsep=0pt\colorbox[HTML]{6CA2E3}{.}}

\textsf{\textbf{Policy}}\quad Policy routes property found in {\fboxsep=0pt\colorbox[HTML]{B5D0F1}{transit}} vehicles to the {\fboxsep=0pt\colorbox[HTML]{91B9EA}{Transit}} {\fboxsep=0pt\colorbox[HTML]{7DADE6}{Dep}}ot {\fboxsep=0pt\colorbox[HTML]{9DC1EC}{V}}ault{\fboxsep=0pt\colorbox[HTML]{A8C8EE}{,}} regardless of where it was {\fboxsep=0pt\colorbox[HTML]{679FE2}{handed}} in{\fboxsep=0pt\colorbox[HTML]{85B2E8}{.}}

\endgroup

\paragraph{ProofWriter.} 208 tokens.
The model chooses option 1, which is \textbf{correct}. Question: Statement: "Gary is cold." Based only on the theory, is the statement true, false, or unknown?

\begin{tabular}{@{}rrp{0.7\linewidth}l@{}}
\toprule
$k$ & Probability & Option & \\
\midrule
1 & 83\% & True: the theory implies the statement. & gold, chosen \\
2 & 0\% & False: the theory implies that the statement is false. &  \\
3 & 17\% & Unknown: the theory implies neither the statement nor its negation. &  \\
\bottomrule
\end{tabular}

\begingroup\scriptsize\setlength{\parindent}{0pt}\setlength{\parskip}{0.3em}\raggedright
\textsf{\textbf{Theory}}\quad Charlie is blue. Charlie is quiet. Dave is blue. Dave is cold. Dave is young. Erin is blue. Erin is quiet. Erin is white. Erin is young{\fboxsep=0pt\colorbox[HTML]{8EB7E9}{.}} {\fboxsep=0pt\colorbox[HTML]{69A0E2}{\underline{Gary}}} {\fboxsep=0pt\colorbox[HTML]{6DA2E3}{\underline{is}}} {\fboxsep=0pt\colorbox[HTML]{83B0E7}{\underline{cold}}}{\fboxsep=0pt\colorbox[HTML]{8DB6E9}{\underline{.}}} Gary is {\fboxsep=0pt\colorbox[HTML]{A9C9EE}{quiet}}. Gary is {\fboxsep=0pt\colorbox[HTML]{C2D9F3}{young}}. {\fboxsep=0pt\colorbox[HTML]{B3CFF0}{All}} {\fboxsep=0pt\colorbox[HTML]{95BCEB}{quiet}} {\fboxsep=0pt\colorbox[HTML]{9CC0EC}{things}} {\fboxsep=0pt\colorbox[HTML]{B3CFF0}{are}} {\fboxsep=0pt\colorbox[HTML]{96BDEB}{furry}}. If Charlie is cold then Charlie is young. {\fboxsep=0pt\colorbox[HTML]{75A8E5}{White}} {\fboxsep=0pt\colorbox[HTML]{94BBEA}{things}} {\fboxsep=0pt\colorbox[HTML]{8FB8E9}{are}} cold. {\fboxsep=0pt\colorbox[HTML]{B8D2F1}{All}} {\fboxsep=0pt\colorbox[HTML]{B5D0F1}{young}} {\fboxsep=0pt\colorbox[HTML]{9CC0EC}{things}} {\fboxsep=0pt\colorbox[HTML]{B9D2F1}{are}} {\fboxsep=0pt\colorbox[HTML]{C1D8F3}{nice}}. If Gary is nice then Gary is young. If {\fboxsep=0pt\colorbox[HTML]{9FC2EC}{something}} is young then {\fboxsep=0pt\colorbox[HTML]{B8D2F1}{it}} {\fboxsep=0pt\colorbox[HTML]{A5C6EE}{is}} nice. If something is {\fboxsep=0pt\colorbox[HTML]{AFCDF0}{furry}} and blue then {\fboxsep=0pt\colorbox[HTML]{97BDEB}{it}} is white. If {\fboxsep=0pt\colorbox[HTML]{C0D7F3}{something}} is {\fboxsep=0pt\colorbox[HTML]{C2D9F3}{young}} and {\fboxsep=0pt\colorbox[HTML]{C4DAF4}{furry}} then it is nice.

\endgroup

\end{document}